# Is it conceivable that neurogenesis, neural Darwinism, and species evolution could all serve as inspiration for the creation of evolutionary deep neural networks?

Mohammed Al-Rawi,

Computer Vision Centre[1],

Universidad Autónoma de Barcelona, Spain

al-rawi@ua.pt

**Abstract**

Deep Neural Networks (DNNs) are built using artificial neural networks. They are part of machine learning methods that are capable of learning from data that have been used in a wide range of applications. DNNs are mainly handcrafted and they usually contain numerous layers. Research frontier has emerged that concerns automated construction of DNNs via evolutionary algorithms. This paper emphasizes the importance of what we call two-dimensional brain evolution and how it can inspire two dimensional DNN evolutionary modeling. We also highlight the connection between the dropout method which is widely-used in regularizing DNNs and neurogenesis of the brain, and how these concepts could benefit DNNs evolution. The paper concludes with a few recommendations for enhancing the automatic construction of DNNs.

---

[1] I wrote this paper in 2018 while I was a research fellow at the Computer Vision Centre, Universidad Autónoma de Barcelona. The effort was abandoned (fortunately, it was saved on Google Drive) since it did not correspond with the group's line of research (Text recognition, word spotting, all about text and document processing). My plan back then was to conduct research on the concept of two-dimensional evolution, both theoretically and practically. However, the paper was completely forgotten as I began working on textual images and languages. Last week I thought of dropping some of my unpublished papers at the Arxiv, and this was one. Disclaimer: I'm not sure if the two-dimensional evolution of DNNs has been used, tested, or applied after 2018.. I never looked at the literature before dropping the paper at the Arxiv. I am currently a Postdoc researcher at DETI, IEETA, the University of Aveiro, Portugal. The paper may have a few linguistic flaws, and I regret not having the time to examine or proofread it. Hence, it is the same copy I wrote four to five years ago.

# 1. Two-Dimensional Brain Evolution

The human brain is the crown jewel of nature. It needed almost four billion years to evolve since the first living cell appeared on earth. There is evidence that brain evolution is two dimensional; the first is the evolution due survivor of the fittest between different species according to Charles Darwin theory of evolution, and the second, the evolution that happens inside the brain itself as claimed in neural Darwinism [1-3]. There are also other theories from neurogenesis stating that neurons migrate to a place in the brain where they will do their work, and differentiation is where a neuron can perform its job depending on its location [4]. The concept thus involves survival, maturation, and integration of newborn neurons, which is controlled in adult neurogenesis by programmed cell death [5]. A neuron, thus, could die during migration, differentiation, or for other unknown reasons. Moreover, circuit remodeling could occur by synaptic redistribution [6]. The dramatic increase in the number of new neurons does not alter the density of synapses in the molecular layer, suggesting thus a readjustment of synaptic connections will occur [7]. Here's a nice quote from [8] that briefly describes neurogenesis:

> *"Neurogenesis, the creation of new brain cells called neurons, occurs primarily before birth. However, a region of the brain called the dentate gyrus, which is involved in memory, continues to produce new neurons throughout life. Recent studies suggest that adding neurons to the dentate gyrus helps the brain to distinguish between similar sights, sounds and smells. This in turn makes it easier to encode similar experiences as distinct memories."*

Another concept/theory that revolves around neurogenesis and neuroplasticity is known as Neural Darwinism Theory that suggests that any neuron that isn't 'fired-and-wired' together into a network is likely to be extinguished [4, 5]. Neural Darwinism was proposed by Gerald Edelman, who was awarded the Nobel Prize for physiology and medicine in 1971. The work presented in [4], however, demonstrated that newborn neurons play a role in expediting this process by "winning out" in a Darwinian manner, i.e., via survival of the fittest type of neuronal battle against other neurons. Here is a quote from Dr. Gerald Edelman [4] on this issue:

> *"Competition for advantage in the environment enhances the spread and strength of certain synapses, or neural connections, according to the "value" previously decided by evolutionary survival. The amount of variance in this neural circuitry is very large. Certain circuits get selected over others because they fit better with whatever is being presented by the environment"*

Some argue that the process may happen by selectively eliminating some synapses rather than eliminating the entire cell/neuron [9]{page 1971, Selective neuron death as a possible memory mechanism Nature}. Dead cells are efficiently removed by immune cells such as microglia in the brain, and immature neuroblasts may also contribute to the clearance of dead cells in the dentate gyrus [5]. That said, and supported by evidence from the two-dimensional brain evolution in species, using a two-dimensional evolution on DNNs might lead to better models.

## 2. Dropout, Neurogenesis, and the One-Dimensional DNNs evolution

The dropout method has been introduced as a simple but effective regularization method aimed at enhancing generalization [10]. The method, which has received nontrivial application in DNNs, is based on randomly dropping units along with their connections from the DNN during training, and then, using these dropped-out units at test time. Although choosing which units to be dropped during training is done randomly, the units will be retained with a fixed probability $p$ independent of the other units. Furthermore, the unit that has been retained with probability $p$ during training will be used in the testing after multiplying its outgoing synaptic weights by $p$. Hence by dropping out some units, they are temporarily removed from the network, along with all their incoming and outgoing connections. Dropout, which is simple and efficient, improves the performance of DNNs and is currently present as a module in several deep learning frameworks, such as TensorFlow, PyTorch, and MXNet.

Hence, then, how can the dropout method be justified? And, can this justification be plausible to the mammalian brain, and more specifically, the human brain? In a comparative sense, Nitish et al in [10] made a nice conjecture between dropout and sexual reproduction as:

> "*A motivation for dropout comes from a theory of the role of sex in evolution (Livnat et al., [11]). Sexual reproduction involves taking half the genes of one parent and half of the other, adding a very small amount of random mutation, and combining them to produce an offspring. The asexual alternative is to create an offspring with a slightly mutated copy of the parent's genes. It seems plausible that asexual reproduction should be a better way to optimize individual fitness because a good set of genes that have come to work well together can be passed on directly to the offspring. On the other hand, sexual reproduction is likely to break up these co-adapted sets of genes, especially if these sets are large and, intuitively, this should decrease the fitness of organisms that have already evolved complicated co-adaptations. However, sexual reproduction is the way most advanced organisms evolved*."

While the aforementioned motivation / inspiration is interesting, another justification for the success and the workability of the dropout method can be derived from neurogenesis, i.e. the birth and death of neurons, and neuroplasticity, i.e. the malleability of neural circuits. Drawing a link between dropout and the way we learn, could it be that our brains use some kind of dropout techniques to enable us learn better, i.e. by shutting a few neurons off during learning and then turning them on some time later? Or, it could be that the dropout method is an ill-posed technique compared to what is really happening in the human brain according to neurogenesis and Neural Darwinism. In reality, what the dropout method is undertaking is perturbing the learning process in a way that part of the units gets less knowledge (in a stochastic shutdown manner) than the other units, and then, using these dropped-out units at test time. With such an approach, and amazingly, the dropout method helps attain better validation/test results.

## 3. Evolutionary DNNs

Astonishingly, there are quite a few DNN evolution research works inspired from Darwinian evolution. For example, the work in [12] led by Google Brain project, which has shown great results compared to handcrafted state-of-the-art DNNs, finds a DNN by running several DNNs on different computers, then choosing the survivor DNN that has the best fit over several generations, where mutations and other evolution parameters have been considered. The work in [12] proposed the so-called Large-Scale Evolutionary Image Classifiers (LSEIC). Another interesting work in [13] proposed the so called Evolutionary Deep Networks for Efficient Machine Learning (EDEN) has shown less success than [12], although their method needed far less operations and can be implemented on a single GPU. These methods, however, use only one dimensional evolution by keeping the fittest DNN and killing the other, which is similar to Charles Darwin Evolution of species. Such a work scenario might start, in the first phase, with a population of different DNN models/algorithms each distributed on a single computer (or worker, as referred to in [12]), select the fittest DNN, then apply mutation and crossover operations on the fittest DNNs to generate an offspring (child-DNNs), and then, train the child-DNN via backpropagation and/or some other training method for a few epochs. Each child-DNN is sent then to the population, letting this DNN population evolve for a few generations to generate an offspring that is better than their ancestors. Backpropagation will have an effect similar to the normal way people learn things, while DNN evolution is intended to cope with Darwinian Evolution, as depicted in Fig. 1. Google Brain researchers in [12], nevertheless, were able to reach an accuracy of 94.6% on CIFAR10 by performing (one-dimensionally) evolution by starting from a simple, one-layer, a task that needed 9×1019 FLOPs (Floating-Point Operations) machine power. The brilliant method, however, may occasionally get trapped in a local optimum that the authors relate to the fixed metaparameters that they used in their model, which is common in evolutionary algorithms.

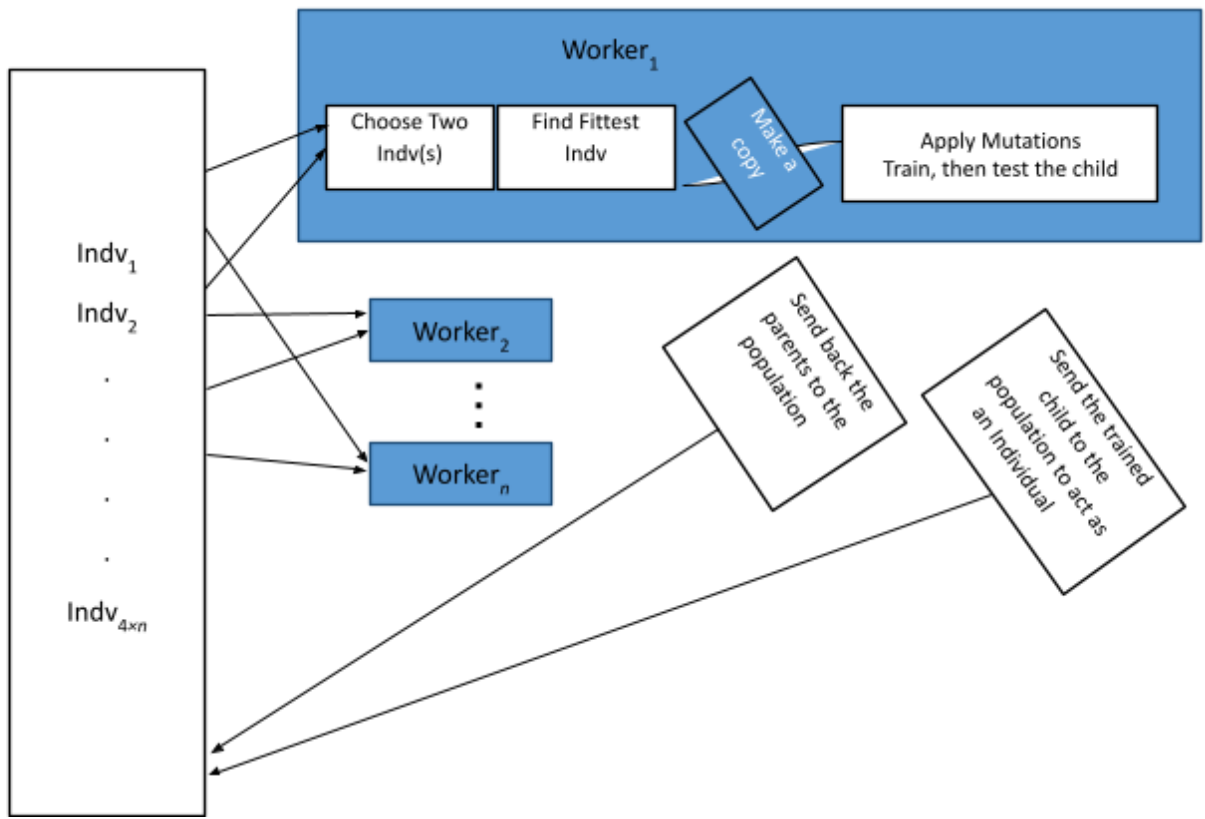

Fig. 1. LSEIC [12] evolutionary classifier has only one dimensional evolution component that follows Charles Darwin species evolution, but has neither neural Darwinism nor neurogenesis components. A 'Worker' is a computer that does all the computations. Worker$_2$, ..., Worker$_n$ has the same components of Worker$_1$ but are listed in a compact form for simplicity. All 'Workers' work in parallel and each individual (Indv) refers to a single DNN structure.

An unpublished work [14], which has been done by some of the authors of [12], reached for the first time state-of-the-art image classifiers using evolutionary algorithms. Their methods, which are based on Regularized Evolution and made use of a VAriant of Tournament selection evolutionary algorithm (REVAT). It is worth mentioning that tournament selection has also been used in [13]. The regularization technique employed in [14] removes the oldest model from a population undergoing tournament selection, and not the worse model as in [12]. A comparison between the three methods is shown in Table 1.

Table 1. Comparison between EDEN, LSEIC and REVAT

| | LSEIC [12] | EDEN [13] | REVAT [14] |
|---|---|---|---|
| Computations | $9\times10^{19}$, using 250 parallel computers | Far Less than $9\times10^{19}$ (as claimed) | Less than 24 hours using Google's TPU** |
| Accuracy on CIFAR-10 | 94.6% | 74.5% | 97.87% |
| Use Dropout | No | Yes | No |
| Start with a small feedforward NN | Yes | NA | NA |
| Accuracy at start, first generation | 21.6% | NA | NA |
| Network Parameters | 5.4M | 0.172M* | 34.9M[¥] |

* The exact number is 173,767, approximated for comparison purposes.
** Google TPU: These revolutionary TPUs were designed to accelerate machine learning workloads with TensorFlow. Each Cloud TPU provides up to 180 teraflops of performance.
[¥] Another more efficient structure has been found with 97.45% accuracy and 2.8M parameters.

## 4. Conclusions

Despite the huge research in neurogenesis and other neuroscience related areas, little knowledge from the brain's biological dynamics has yet contributed to DNNs evolution. To add to the discussion, would a method inspired from what is known from neurogenesis give further enhancement than dropout? Such a method would include, randomly adding new born units (that are randomly initialized after birth), migrating some other units from one position/layer to another, killing some other units, or even migrating some convolution filters, from one position/layer to another, inter and/or intra layer migration, removing some synaptic connections, etc. Such an approach can be called Evolutionary-Rewired-DNNs, analogous to how neurogenesis, neuroplasticity and stochastic and Darwinian contexts are rewiring the brain. Part of these operations were the focus of LSEIC [12], EDEN [13] and REVAT [14], but across population evolution. It is worth mentioning that even the so-called neuroevolution works, such as that of [15] and the recent ones, are across population evolution "Neuroevolution starts out with a uniform population of networks with zero hidden nodes: page 111, Section 3.4 of Ref [15]". Thus, the foreseen approach we are suggesting should contain across population evolution (Darwinian Evolution), within individual evolution (Neural Darwinism), and within individual / structure training (via Backpropagation). The within individual evolution (i.e. Neural Darwinism) could be implemented by stochastically removing the bad neurons and or synapses (filters, layers, etc.), as well as introducing new ones based on neurogenesis. The answers to the above raised questions, and the suggested models, can be resolved by someone who affords a powerful 9X10$^{19}$ FLOPs machine. Nonetheless, the aforementioned discussion on neurogenesis and Neural Darwinism indicate that further development will greatly improve the development of evolutionary DNNs, and thus, further evolutionary models should be designed based on evidence from neurogenesis and Neural Darwinism, in addition to across structure evolution.

The human brain, which is an extremely complex structure, can 'quickly' learn very simple things; For example, adding two one-digit numbers, in addition to 'slowly' and spontaneously learning complex things. The brain has acquired its learning capability through a long time evolution cycle. And thus, a newborn is already equipped, via inheritance, with a highly intelligent brain that will be pruned during his / her lifetime. That said, it is unclear whether the wide-spread complex deep neural networks can learn simpler tasks than those that they have been originally designed for. Based on this notion, it would be highly beneficial if researchers develop and/or generate universal or general purpose DNNs rather than customized via evolutionary methods instead of several ones that perform well on the validation set for a specific problem, although their generalization might still be under question. Hopefully, the day that an inherited deep neural network can learn simple and complex tasks is near.

.